\documentclass[preprint,12pt]{elsarticle}

\usepackage{mathtools}

\usepackage{setspace}
\usepackage{amssymb, amsmath, graphicx, subfigure,multirow, color, rotating,verbatim, comment,hyperref}
\usepackage[table,xcdraw]{xcolor}
\usepackage[justification=centering]{caption}
\journal{Neurocomputing}

\usepackage{array}
\usepackage{float}
\newcolumntype{L}{>{\centering\arraybackslash}m{1.5cm}}
\newcolumntype{M}{>{\centering\arraybackslash}m{1.1cm}}
\usepackage{url}
\usepackage{tabularx,booktabs}

\usepackage{multirow}
\usepackage{graphicx}

\begin{document}

\begin{abstract}
k-nearest neighbour (kNN) is one of the most prominent, simple and basic algorithm used in machine learning and data mining. However, kNN has limited prediction ability, i.e., kNN cannot predict any instance correctly if it does not belong to any of the predefined classes in the training data set. The purpose of this paper is to suggest an Advanced kNN (A-kNN) algorithm that will be able to classify an instance as \textit{unknown}, after verifying that it does not belong to any of the predefined classes. Performance of kNN and A-kNN is compared on three different data sets namely iris plant data set, BUPA liver disorder data set, and Alpha Beta detection data set. Results of A-kNN are significantly accurate for detecting unknown instances. 

\end{abstract}


\begin{frontmatter}
\title{Advanced kNN: A Mature Machine Learning Series}

\author[cs]{Muhammad Asim}
\author[cs1]{Muaaz Zakria}

\address[cs]{Department of Electrical Engineering, Riphah International University, Lahore, Pakistan}
\address[cs1]{Information Technology University (ITU)-Punjab, Lahore, Pakistan}
\end{frontmatter}

\section{Introduction}
k-nearest neighbour is among the top 10 classification algorithms as described by \cite{wu2008top}, \cite{zhang2018novel}, and \cite{zhu2017local} used in KDD~\footnote{KDD: Knowledge Discovery in Databases (KDD) is the process of discovering useful knowledge from a collection of data.} and data mining. kNN algorithm is a simple but effective non-parametric classification method \cite{hand2001principles}. 
The principle behind Nearest Neighbour Classification is quite elementary. The instances are categorized based on the class of their nearest neighbours. It is often useful to take more than one neighbours into account so the technique is more commonly referred as k-Nearest Neighbour classification where k nearest neighbours are utilized in determining the class \cite{cunningham2007k}. 
It is called a lazy learner, as it performs classification tasks without  building its model ~\cite{ref:zhang2019cost}. Model-based classifiers are those which builds a model and learns  from given and existing training data set. Then using that model its accuracy is tested with test data set. The same model is used to predict the label~\footnote{In this paper, label and class is used interchangeably} of new instances. Unlike these model-based classifiers, kNN keeps all previous examples in memory. So for new prediction, kNN finds k nearest instances and decides class of new instance to be the class same as of neighbors. 
Imbalanced~\footnote{Data set having different number of instances in positive and negative class.} data set also has some impacts on classifier performance. 
kNN is a type of supervised machine learning (ML) algorithm which can be used for both classification as well as regression predictive problems. However, it is mainly used for classification. The main purpose of kNN is to estimate the classification of an unseen instance using the class label of the instance or instances that are nearest to it \cite{bramer2007principles}. 

  Besides various improvements over the years in kNN, it still has limited capacity of prediction. kNN cannot predict any instance correctly if it does not belong to any of the predefined classes in the training data set. As an example,  if kNN is trained with a data set of patients who are either healthy or suffering from cancer, and in test set there is a patient of diabetes. The traditional kNN algorithm will classify that diabetes patient as a healthy or a cancer patient as it has only two options, while both the options are wrong. 

To overcome this issue, Advanced kNN (A-kNN) will identify it as an \textit{unknown} instance and will inform the domain expert about it. Domain expert will guide the algorithm on this instance, whether to add it as separate class to training set or classify it as one of the predefined classes.
In this way, by employing concepts of \textit{Training Class Area}, \textit{Area of class} and \textit{Gap Constant}, A-kNN will be quite mature to identify \textit{unknown} instances without requiring any predefined data in the training set. All these concepts are explored in coming sections.

Section \ref{sec:related_work} will cover related work with respect to improvement in kNN algorithm. We introduce our Advanced kNN algorithm in section \ref{sec:aknn}. Section \ref{sec:experiment} will cover the experiments and results and in the last section we will discuss the conclusion and directions for future research.

\section{Related work}\label{sec:related_work}

kNN methodology is still a hot topic for research in machine learning and data mining despite the fact that the algorithm was first proposed in 1967. The kNN rule as described by Devijver and Kittler, has been widely used since it is effective, when probability distributions of the feature variables are not known  \cite{devijver1982pattern}. Various researchers have proposed various methods to improve this algorithm. Hart proposed a computationally simple local search method as Condensed Nearest Neighbour (CNN) by minimizing the number of stored patterns and storing only a subset of the training set for classification  \cite{hart1968condensed}. The basic idea is that patterns in the training set may be very similar and many of the instances do not add extra information and thus may be discarded. Gates proposed the Reduced Nearest Neighbour (RNN) rule that aims to further reduce the stored subset after having applied CNN \cite{gates1972reduced}. It simply removes those elements from the subset which will not cause an error. 

Li et al. suggested an improved kNN algorithm for text categorization. Their main  focus is on the selection of the parameter k. In traditional kNN algorithms, the value of k is already fixed. If k is very large, then the big classes will overcome the small ones. Practically, the value of k is usually optimized by running various trials on the training and validation sets. However, this method is not suitable in those cases where there is no option to perform cross-validation, like in online classification. To tackle this problem, they proposed a revised k-Nearest Neighbor algorithm, which employs various values of k for various classes, rather than a fixed k value  \cite{li2003improved}.
Zhang et al. proposed a computation method for k-parameter for the purpose of kNN approximate prediction based on Sparse learning \cite{cheng2014knn}, known as S-kNN \cite{zhang2018novel}. Their proposed S-KNN algorithm works out an optimized value of k for each and every test sample. This implies that the value of k can be different for various test samples. 
Similarly Zhang et al. put forward a new kNN method to learn different values of k for different instances of test data by following the distribution of training data. They named it as  Correlation Matrix kNN (CM-kNN) \cite{zhang2017learning}. This method involves usage of previous knowledge inherent in training data, including the correlation among data points, the removal of noisy data, and preserving the local structures of the data.

Liu et. al also proposed a new method unlike traditional kNN methods and named it Mutual Nearest Neighbours (MNN)  \cite{liu2010new}. It makes use of the concept of mutual nearest neighbor of the unknown instance to determine its label. To approximate the label of that instance, MNN first identifies its mutual nearest neighbors and then goes on to make a decision. Their claim is that the predicted label is more creditable as it is derived from the intimate neighbors. Moreover, some outliers will also be left out. Toyama et. al, by utilizing the marginal distribution of kth nearest neighbors in low dimensions, put forward a probably correct approach of kNN, which is a probabilistic modification of the partial distance searching  \cite{toyama2010probably}. Guo et. al formalized a method which he called MkNN to automatically determine a proper value of k for different datasets. It is quite similar to the cross-validation technique. MkNN firstly develops a kNN model for the data, and then chooses an optimal value of k on the basis of its classification accuracy \cite{guo2003knn}.

\section{Advanced kNN} \label{sec:aknn}
Before going through A-kNN, we will first sift through the original kNN algorithm.

\subsection{kNN Algorithm}
The following steps depict the working of kNN algorithm\cite{tpoint}:
\begin{enumerate}

\item[1] Collect training data set.
\item[2] Set integer value of k.
\item[3] To predict label of new instance/point,
\begin{enumerate}
\item[3.1] Calculate the distance between new instance and all instances of training data set. 
\item[3.2] Sort list of distances in ascending order.

\item[3.3] Select only top  k (lowest) distances and their class labels.

\item[3.4] Class of the new instance will be the most frequent class of top k selected instances. 

\end{enumerate}
\item[4] End.
\end{enumerate}
\subsection{Improvement in traditional kNN}
Till now, the bulk of research on kNN has been revolving around two areas. One is to select the best value of k, and the other is to select distance calculating method. Commonly used values of k are 1, 3, 5, 7, and so on. While the most used methods for distance calculation are following:
\begin{enumerate}
\item Euclidean Distance:
\begin{equation}
d(Euclidean) = \sqrt{\sum_{i=1}^{k} (x_i-y_i)^2} 
\end{equation}
\item Manhattan Distance:
\begin{equation}
d(Manhattan) = \sum_{i=1}^{k}|x_i-y_i|
\end{equation}
\item Minkowski Distance:
\begin{equation}
d(Minkowski) = (\sum_{i=1}^{k}(|x_i-y_i|)^q))^{\dfrac{1}{q}}
\end{equation}
\end{enumerate}

\subsection{Advanced kNN}

As mentioned previously, our Advanced kNN will make use of a \textit{Training Class Area (TCA)}, \textit{Area of Class}  and  \textit{Gap Constant} to classify whether an instance belongs to some unknown class or not. Following steps show the working of Advanced kNN algorithm:

\begin{enumerate}

\item[1] Collect training data set.
\item[2] Set integer value of k.
\item[3] Define TCA i.e., the training class area of each class (will be explained next). 
\item[4] To predict label of new instance/point,
\begin{enumerate}
\item[4.1] Calculate the distance between new instance and all instances of training data set. 
\item[4.2] Sort list of distances in ascending order.

\item[4.3] Select only top k distances and their class labels. Smallest distances is called $min\_dist$ .

\item[4.4] Find most frequently occurring class in the list of top k instances called \textbf{expected class}. 
\begin{enumerate}
\item[]if. ( $min\_dis t>$ area of expected class)

\item[] then. Class of new instance is\textit{ unknown}.

\item[]else. Class of new instance is label of expected class. 
\end{enumerate} 

\end{enumerate}
\item[5] End.
\end{enumerate}
\subsubsection{Defining Area of a class}
Defining an area for a class has no specific rules or an equation as it is a new area of research in A-kNN. Basically defining area of class majorly depends on the nature of the data. In this research we defined the area as given below:
\begin{enumerate}
\item[1.] Select one of the above distance calculation formula.
\item[2.] Using that formula calculate distance of all instances in the training set with each other.
\item[3.] Sort it in descending order and select maximum distance. 
\item[4.] This distance will be called \textit{Training Class Area} (TCA).
\item[5.] Using this TCA, domain expert will decide boundary of a class based on the nature of whole data. If classes are closer to each other, boundary will be narrow about 1.5 times of TCA or 2 times of TCA. As the distance among classes increases, gap can be widened and area of class can be decided to be about gc times of TCA, where gc is called Gap constant.  Area of class can be more widened or tied nearer, depending upon the nature of data.
\end{enumerate}

\begin{table}[htbp]
  \centering
  \caption{Data set for analysis}
    \begin{tabular}{|c|c|c|l|c|}
    \hline
    \multicolumn{1}{|p{4.785em}|}{Employ} &\multicolumn{1}{|p{4.785em}|}{Salary(\$)} & \multicolumn{1}{p{4.785em}|}{Scale} & \multicolumn{1}{p{5.855em}|}{Original   $   $Title} & \multicolumn{1}{p{4.93em}|}{Assigned Label} \\
    \hline
    E1 & 581   & 17    & Gazetted & G \\
    \hline
    E2 & 710   & 18    & Gazetted & G \\
    \hline
    E3 & 370   & 15    & Non-Gazetted & N \\
    \hline
    E4 & 413   & 16    & Non-Gazetted & N \\
    \hline
    E5 & 329   & 16    & Non-Gazetted & ? \\
    \hline
    E6 & 626   & 18    & Gazetted & ? \\
    \hline
    E7 & 129   & 4     & Class-4 & ? \\
    \hline
    E8 & 968   & 21    & Bureaucrat & ? \\
    \hline
    \end{tabular}%
  \label{tab:dataset}%
\end{table}%

\begin{figure}[ht]
\begin{center}
\caption{Data points}
\includegraphics[scale=1.3]{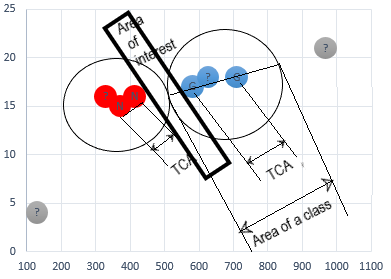} 
\label{pic:datapoints}

\end{center}
\end{figure}

Before moving on to broader experimentation, we have a dataset of simple eight points given in table \ref{tab:dataset} and visualized in figure \ref{pic:datapoints}. Salary (in \$) and employee scale are the two features which will be used to predict the title  or label~\footnote{Here the words `title' and `label' are used in the same sense, title of last two employees is just for ease of the reader.} of employee. Scale 15, and 16 represent non gazetted officers while scale 17 and 18 are called gazetted officers. Data of first four employs $E1-E4$ is used for training. While label of next four $E5-E8$ is to be predicted. So using simple kNN and setting k=1 prediction of E5 and E6 is very easy and obvious. E5 will be predicted as Non-gazetted (N), while label of E6 will be predicted as gazetted (G). E7 and E8 are the points of our interest. As the actual title of E7 is \textit{Class-4} while that of E8  is \textit{Bureaucrat}. However, there is no data of both of these titles in our training set. So according to kNN, E7 is near to Non-gazetted officer while E8 is near to gazetted officer. So kNN will classify E7 as N, while E8 as G, which are actually wrong. What will A-kNN do in this case? If there is point somewhere outside from the regions of already available classes in training data set, it will be classified as an unknown category. So a domain expert will be notified to guide A-kNN for this point. 

In figure \ref{pic:datapoints}, the main area of interest in previous literature is the bold lined rectangle. To decide whether the instances in the area of bold rectangle will belong to G class or N. Here in this research we are interested in the instance beyond this area, which lies very far from all previous class like E7 and E8. Distance between E1 and E2 is marked as TCA for Gazetted class, while distance between E3 and E4 is marked as TCA for Non-gazetted class. Area of both classes is encircled.

Using euclidean distance on the data set in table \ref{tab:dataset}, TCA of G class is 129, while TCA of N class is 43. Value of gap constant is set to be 1.5. So area  of G class is 193.5 while area of N class is calculated to be 64.37.
Setting k = 1, E5 is very closed to E3. Euclidean distance between these two instances is 41 which is minimum, and lower than the area of N class (64.37), so obviously it will be classified as N. Similarly euclidean between E6 and E1 is 45.01 which is minimum and lower than area of G class (193.5), so its label will be predicted as G. On the other hand E7 is nearest to E3, but euclidean distance between E7 and E3 is 241.25 which very very greater than area of N class (64.37). It means that E7 is very very far from all the classes and will be classified as unknown class. So domain expert should be notified to guide the machine on this instance. Now it is in the hands of domain expert whether he classifies and add this instance to nearest class (Non-gazetted) or labels it as a new class (class-4), and retrain the machine. Similarly E8 is nearest to E2, their euclidean distance is 258$>>$193.5 (area of G class). So domain expert will be requested for guidance on this instance too.   

\subsubsection{Impact of Gap Constant}

\begin{figure}[ht]
\begin{center}
\caption{Impact of Gap constant}
\includegraphics[scale=0.65]{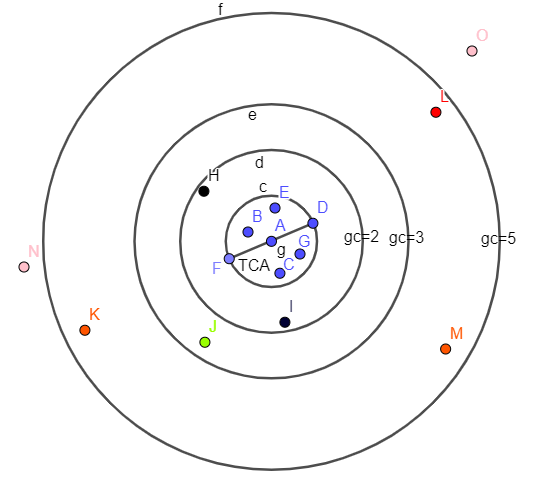} 
\label{pic:gcImpact}
\end{center}
\end{figure}

Figure \ref{pic:gcImpact} depicts the impact of gap constant on area of class. Initially the training class only contains blue points (A, B, C, D, E, F and G). Maximum distance of this class is between point D and point F, so circle of diameter equal to maximum distance (i.e.inner circle) is TCA. When gc=1, all the points except blue are out of this class. For gc=2, area of class expands and black points (i.e. H and I) also become members of this class. J may get membership of this class by expanding area of class more to circle e with gc=3. Points N and O are even not members of this class with gc=5. So as we increase the gc, area of the class increases, and far more points will be classified as a members of this class.    

\section{Experiments and Results}\label{sec:experiment}
\begin{table}[htbp]
  \centering
  \caption{Data sets details}
    \begin{tabular}{|p{4.715em}|c|c|c|p{5.645em}|}
    \hline
    Data set Title & \multicolumn{1}{p{5em}|}{Number of Instances} & \multicolumn{1}{p{4.785em}|}{Number of Attributes} & \multicolumn{1}{p{4.5em}|}{Number of Classes} & Class Names \\
    \hline
    Iris Plants Database & 150   & 4     & 3     &  Setosa, $  $Versicolour, $  $Virginica \\
    \hline
    BUPA $  $liver $  $disorders & 345   & 6     & 2     & 1, 2 \\
    \hline
    Alpha Beta Detection & 199   & 2     & 2     &  0, 1 \\
    \hline
    \end{tabular}%
  \label{tab:ucidatasets}%
\end{table}%

We have carried out experiments to evaluate the performance of A-kNN in comparison with the traditional kNN algorithm. Experiments have been performed on data sets downloaded from UCI repository. The details of data sets are in given table \ref{tab:ucidatasets}. Both A-kNN and traditional kNN have been implemented in python.
\subsection{Experimental procedure}
Data sets are split into training and testing set in ratio of 70:30. Both A-kNN and traditional kNN are trained on the training set. Performances of both algorithms are evaluated in two ways. First both are evaluated using original test set, which means all the data in the test set belongs to one of the class and no unknown (vary far from seen class) data is fed.  In this case the performances of both A-kNN and traditional kNN is comparable. Now, instances of unknown data are added to the test set. These instances are very far from all previous classes, which means all these instances must not belong to any of the previous class and the algorithm should indicate them as unknown instances. So that the domain expert may take necessary steps about this data. Here the importance of A-kNN will be highlighted. Because traditional kNN will not be able to indicate unknown data, but it will classify these instances as one of the previous classes, based on nearest neighbors. 
\subsection{Experiments on Iris Plants Data Set}

\begin{table}[]
\caption{Results of Iris data set with kNN and A-kNN}
\resizebox{\textwidth}{!}{%
\begin{tabular}{|p{1.5cm}|p{1.5cm}|p{1.5cm}|p{2cm}|p{2cm}|p{2cm}|p{2cm}|p{2cm}|p{2cm}|}
\hline
\multirow{2}{*}{\textbf{K}} & \multirow{2}{*}{} & \multirow{2}{*}{\textbf{gc}} & \multicolumn{2}{c|}{\textbf{Accuracy}} & \multicolumn{4}{c|}{\textbf{Misclassified unknown instances (out of 20)}} \\ \cline{4-9} 
 &  &  & \textbf{Without unknown instances} & \textbf{With unknown instances} & \textbf{Total} & \textbf{classified as label setosa} & \textbf{classified as label virsicolor} & \textbf{classified as label verginica} \\ \hline
\textbf{1} & \textbf{KNN} & \textbf{-} & \textbf{0.98} & \textbf{0.68} & \textbf{20} & \textbf{4} & \textbf{0} & \textbf{16} \\ \hline
1 & AKNN & 1 & 1.00 & 1.00 & 0 & 0 & 0 & 0 \\ \hline
1 & AKNN & 1.5 & 1.00 & 1.00 & 0 & 0 & 0 & 0 \\ \hline
1 & AKNN & 2 & 1.00 & 0.98 & 1 & 1 & 0 & 0 \\ \hline
1 & AKNN & 5 & 1.00 & 0.94 & 4 & 4 & 0 & 0 \\ \hline
1 & AKNN & 10 & 1.00 & 0.92 & 5 & 4 & 0 & 1 \\ \hline
1 & AKNN & 100 & 1.00 & 0.75 & 16 & 6 & 0 & 10 \\ \hline
1 & AKNN & 1000 & 1.00 & 0.72 & 18 & 6 & 0 & 12 \\ \hline
\textbf{7} & \textbf{KNN} & \textbf{-} & \textbf{0.98} & \textbf{0.68} & \textbf{20} & \textbf{4} & \textbf{0} & \textbf{16} \\ \hline
7 & AKNN & 1 & 0.98 & 0.97 & 0 & 0 & 0 & 1 \\ \hline
7 & AKNN & 1.5 & 0.98 & 0.97 & 0 & 0 & 0 & 1 \\ \hline
7 & AKNN & 2 & 0.98 & 0.95 & 1 & 1 & 0 & 1 \\ \hline
7 & AKNN & 5 & 0.98 & 0.91 & 5 & 4 & 0 & 1 \\ \hline
7 & AKNN & 10 & 0.98 & 0.89 & 6 & 4 & 0 & 2 \\ \hline
7 & AKNN & 100 & 0.98 & 0.74 & 16 & 6 & 0 & 10 \\ \hline
7 & AKNN & 1000 & 0.98 & 0.71 & 18 & 6 & 0 & 12 \\ \hline
\end{tabular}%
}
\label{tab:irisresults}%
\end{table}

Details of iris plants database is given in table \ref{tab:ucidatasets}. Total instances in iris data set are 150. By keeping 70:30 ratio, number of instances in training set is 105 while number of instances in original test set becomes 45.
Number of unknown instances is 20. For k=1, 7 and using the test set without unknown instances, the results of KNN and A-KNN comes out to be similar. However, with unknown instances, it can be seen that results are highly different. KNN misclassified all of the unknown instances as one of the predefined class in training set. Whereas A-KNN discriminates unknown and known instances, and clearly identifies very far instances as unknown instances. Further A-KNN is strongly dependent on the value of gap constant.  As the value of gap constant increases, area of class becomes widen. When any of the instances gets into the area of any class, it is classified as that class. With a gap constant of 1000, A-KNN  miss-classifies 18 unknown instances too.

\subsection{Experiments on BUPA Liver Disorders Data Set}

In this particular data set, the total number of instances we have is 345. The total number of attributes are 6 while the target classes are two namely 1 in case of a liver disorder and 2 in case of absence. When it is divided in the ratio of 70:30, the training set comes out to be of 241 and testing set to be 104. The results are shown in table \ref{tab:bupa}. Unknown instances here are again 20 in number. For k=1 and 7, KNN shows almost dismal results both with and without unknown instances. On the contrary, A-KNN with k=1 outperforms KNN with k=1 \& 7, and A-KNN with k=7. It can be seen that for A-KNN with both k=1 and 7, the number of misclassified instances rise when the value of gap constant is increased beyond 2. 

\begin{table}[]
  \caption{Results of BUPA Liver Disorders Data Set with kNN and A-kNN}
\resizebox{\textwidth}{!}{%
\begin{tabular}{|p{2cm}|p{2cm}|p{2cm}|p{2cm}|p{2cm}|p{3cm}|p{3cm}|p{1cm}|}
\hline
\multirow{2}{*}{\textbf{K}} & \multirow{2}{*}{} & \multirow{2}{*}{\textbf{gc}} & \multicolumn{2}{c|}{\textbf{Accuracy}} & \multicolumn{3}{c|}{\textbf{Misclassified unknown instances (out of 20)}} \\ \cline{4-8} 
 &  &  & \textbf{Without unknown instances} & \textbf{With unknown instances} & \textbf{Total} & \textbf{classified as label 1} & \textbf{classified as label 2} \\ \hline
\textbf{1} & \textbf{KNN} & \textbf{-} & \textbf{0.57} & \textbf{0.48} & \textbf{20} & \textbf{1} & \textbf{19} \\ \hline
1 & AKNN & 1 & 1 & 1 & 0 & 0 & 0 \\ \hline
1 & AKNN & 1.5 & 1 & 0.99 & 1 & 0 & 1 \\ \hline
1 & AKNN & 2 & 1 & 0.99 & 1 & 0 & 1 \\ \hline
1 & AKNN & 5 & 1 & 0.90 & 12 & 1 & 11 \\ \hline
1 & AKNN & 10 & 1 & 0.86 & 17 & 2 & 15 \\ \hline
1 & AKNN & 100 & 1 & 0.83 & 20 & 4 & 16 \\ \hline
1 & AKNN & 1000 & 1 & 0.83 & 20 & 4 & 16 \\ \hline
\textbf{7} & \textbf{KNN} & \textbf{-} & \textbf{0.61} & \textbf{0.51} & \textbf{20} & \textbf{2} & \textbf{18} \\ \hline
7 & AKNN & 1 & 0.73 & 0.77 & 0 & 0 & 0 \\ \hline
7 & AKNN & 1.5 & 0.73 & 0.76 & 1 & 0 & 1 \\ \hline
7 & AKNN & 2 & 0.73 & 0.76 & 1 & 0 & 1 \\ \hline
7 & AKNN & 5 & 0.73 & 0.67 & 12 & 1 & 11 \\ \hline
7 & AKNN & 10 & 0.73 & 0.63 & 17 & 2 & 15 \\ \hline
7 & AKNN & 100 & 0.73 & 0.61 & 20 & 2 & 18 \\ \hline
7 & AKNN & 1000 & 0.73 & 0.61 & 20 & 2 & 18 \\ \hline
\end{tabular}%
}
\label{tab:bupa}%
\end{table}

\subsection{Experiments on Alpha Beta Detection Data Set}
This data set contains a total of 199 instances. Keeping training and testing ratio 70:30, the number of instances in training set are  139 while 60 are in test set. Similar to the previous experiments, 20 unknown instances have been kept here again. The results exhibit that 100\% accuracy is achieved by both KNN and A-KNN employing k=1 and 7 when there are no unknown instances. With the introduction of unknown instances, the accuracy of KNN dips down by 25\%, while that of A-KNN by a mere 7.5\% when the gap constant is kept till 2. As also shown in our previous experimentations, the number of misclassified instances start increasing when we move the gap constant above 2.


\begin{table}[]
\caption{Results of Alpha Beta Detection data set with kNN and A-kNN}
\resizebox{\textwidth}{!}{%
\begin{tabular}{|p{2cm}|p{2cm}|p{2cm}|p{2cm}|p{2cm}|p{3cm}|p{3cm}|p{1cm}|}
\hline

\multirow{2}{*}{\textbf{K}} & \multirow{2}{*}{} & \multirow{2}{*}{\textbf{gc}} & \multicolumn{2}{c|}{\textbf{Accuracy}} & \multicolumn{3}{c|}{\textbf{Misclassified unknown instances  (out of 20)}} \\ \cline{4-8} 
 &  &  & \textbf{Without unknown instances} & \textbf{With unknown instances}& \textbf{Total} & \textbf{classified as label 0} & \textbf{classified as label 1} \\ \hline
\textbf{1} & \textbf{KNN} & \textbf{-} & \textbf{1} & \textbf{0.75} & \textbf{20} & \textbf{14} & \textbf{6} \\ \hline

1 & AKNN & 1 & 1 & 0.925 & 6 & 3 & 3 \\ \hline
1 & AKNN & 1.5 & 1 & 0.925 & 6 & 3 & 3 \\ \hline
1 & AKNN & 2 & 1 & 0.925 & 6 & 3 & 3 \\ \hline
1 & AKNN & 5 & 1 & 0.89 & 9 & 5 & 4 \\ \hline
1 & AKNN & 10 & 1 & 0.83 & 13 & 5 & 8 \\ \hline
1 & AKNN & 100 & 1 & 0.75 & 20 & 12 & 8 \\ \hline
1 & AKNN & 1000 & 1 & 0.75 & 20 & 12 & 8 \\ \hline
\textbf{7} & \textbf{KNN} & \textbf{-} & \textbf{1} & \textbf{0.75} & \textbf{20} & \textbf{14} & \textbf{6} \\ \hline
7 & AKNN & 1 & 1 & 0.925 & 6 & 3 & 3 \\ \hline
7 & AKNN & 1.5 & 1 & 0.925 & 6 & 3 & 3 \\ \hline
7 & AKNN & 2 & 1 & 0.925 & 6 & 3 & 3 \\ \hline
7 & AKNN & 5 & 1 & 0.887 & 9 & 5 & 4 \\ \hline
7 & AKNN & 10 & 1 & 0.837 & 13 & 5 & 8 \\ \hline
7 & AKNN & 100 & 1 & 0.75 & 20 & 12 & 8 \\ \hline
7 & AKNN & 1000 & 1 & 0.75 & 20 & 12 & 8 \\

\hline
\end{tabular}%
}
\label{tab:alpha}%
\end{table}

\newpage

\section{Conclusions and Future Work}\label{sec:conclusions}

The presented results of our experiments clearly demonstrate the utility of our Advanced-KNN algorithm over the traditional KNN when dealing with instances that cannot be categorised under any of the prescribed classes. By varying the gap constant which is linked with the area of class and making use of two values of k, we have observed various scenarios where the unknown instances being categorised under an unknown class. It is also depicted that when the gap constant is increased above 2, the amount of misclassifed unknown instances start increasing. We in our study have striven to offer a new avenue of research. For the days to come, calculating an optimal value of the gap constant (area of the class) for more accurate classifications, is one area which can be researched.

 \newpage 
\bibliography{KNN}

\end{document}